\documentclass{article} 
\usepackage[preprint]{colm2026_conference}

\usepackage{microtype}
\usepackage{hyperref}
\usepackage{url}
\usepackage{booktabs}
\usepackage{multirow}
\usepackage{listings}

\usepackage{amsmath,amsfonts,bm}

\def\eqref#1{equation~\ref{#1}}

\def\1{\bm{1}}

\DeclareMathAlphabet{\mathsfit}{\encodingdefault}{\sfdefault}{m}{sl}
\SetMathAlphabet{\mathsfit}{bold}{\encodingdefault}{\sfdefault}{bx}{n}

\usepackage{lineno}
\usepackage{graphicx}
\usepackage{tablefootnote}
\usepackage{amsmath}
\usepackage{mathtools}
\usepackage{dsfont}

\definecolor{darkblue}{rgb}{0, 0, 0.5}
\hypersetup{colorlinks=true, citecolor=darkblue, linkcolor=darkblue, urlcolor=darkblue}

\title{Policy-Guided Stepwise Model Routing for\\Cost-Effective Reasoning}

\author{Wenwen Si, Insup Lee, and Osbert Bastani\\ 
Department of Computer and Information Science\\
University of Pennsylvania\\
Philadelphia, PA 19104, USA \\
\texttt{wenwens@seas.upenn.edu} 
}

\begin{document}

\ifcolmsubmission
\linenumbers
\fi

\maketitle

\begin{abstract}
Inference-time computation has greatly enhanced the performance of large language models (LLMs) on challenging reasoning tasks, but this strategy can incur high inference costs. One solution is to route intermediate chain-of-thought (CoT) states to language models of different sizes; however, existing approaches rely on handcrafted routing strategies that limit performance, or on training large process reward models that may be infeasible in many applications. We formulate stepwise model routing as a constrained decision-making problem, which we solve by training a small control policy using reinforcement learning in conjunction with threshold calibration to tune the performance-efficiency tradeoff. We validate our method on three math benchmarks (GSM8K, MATH500, and OmniMath) on both open and closed models. Our method consistently improves the accuracy–cost tradeoff compared to handcrafted approaches, while achieving a comparable tradeoff to methods that require training large process reward models.
\end{abstract}

\section{Introduction}

Inference-time scaling techniques, such as Chain-of-Thought (COT)~\citep{wei2022chain}, Self-Consistency~\citep{wang2022self}, Tree Search~\citep{yao2023tree}, and auto-regressive MCMC sampling~\citep{karan2025reasoning}, have substantially improved the capabilities of large language models (LLMs) on reasoning tasks such as mathematics and programming \citep{hestness2017deep, kaplan2020scaling, sardana2023beyond}. At the same time, these methods incur significantly higher inference costs. To address this issue, recent work has leveraged capability--cost tradeoffs among LLMs of different sizes, dynamically routing generation steps to an appropriate size LLM to optimize accuracy under a given computation budget. At each step, the router inspects the current reasoning trace and chooses which size model to use to generate the next unit of output. Routing decisions can occur at the token~\citep{leviathan2023fast, chen2023accelerating}, reasoning step level~\citep{leviathan2023fast, liao2025reward}, or task~\citep{chen2023frugalgpt,ding2024hybridllm,si2025ccpo} levels.

Existing approaches either rely on heavily trained process reward models, such as Reward-Guided Speculative Decoding (RSD)~\citep{liao2025reward}, which require substantial data and compute, or utilize simple uncertainty metrics such as verbal uncertainty~\citep{pan2025specreason} or Gaussian mixture models~\citep{lee2025steer} with threshold calibration via grid search.

We propose a novel approach that relies on data to train an effective decision-making policy without requiring the scale of data needed to finetune an LLM. Specifically, viewing the routing problem as a sequential decision-making problem, we train a control policy to solve this problem; rather than using an LLM as the policy, we use a simple neural network architecture to ensure data efficiency. In addition, a key challenge with existing methods is that targeting a specific performance--cost tradeoff requires hyperparameter tuning. To avoid this issue, our algorithm performs threshold calibration jointly with policy optimization to automatically target a user-selected accuracy level between the accuracies of the large and small LLMs.
In summary, our contributions are:
\begin{itemize}
\item We formulate model routing as a constrained Markov decision process (CMDP).
\item Focusing on guidance at the level of reasoning steps, we propose a novel algorithm for solving our routing CMDP by performing constrained policy learning in conjunction with threshold calibration to target a desired accuracy level.
\item We empirically demonstrate that our method improves the cost-effectiveness of stepwise routing on GSM8K, MATH500, and OmniMath while achieving target accuracies across a range of open and closed models.
\end{itemize}

\section{Related Work}

\textbf{Inference-time scaling.} Sophisticated large language models (LLMs) often adopt structured inference strategies to improve performance or efficiency at test time. Decomposing complex tasks into a sequence of simpler reasoning steps is among the most general methodologies with significant performance improvement, commonly referred to as chain-of-thought (CoT)~\citep{wei2022chain}. Related approaches include Tree of Thoughts~\citep{yao2023tree}, tree search guided by process reward models~\citep{lightman2023let}, and iterative single-model sampling methods based on MCMC-style sampling~\citep{brown2024large, karan2025reasoning}, which sacrifices inference cost and latency for improved performance. In contrast, speculative decoding~\citep{leviathan2023fast, chen2023accelerating} was developed to reduce inference cost while preserving the exact output distribution of the target model. It typically employs a smaller draft model to propose candidate tokens, while the larger model is used to verify those proposals through its conditional probabilities and to correct mismatches when necessary. This makes speculative decoding closer to our target setting. However, its strict unbiasedness requirement, which guarantees that the final token distribution matches that of the large model, also limits flexibility in exploring more diverse completions.

\textbf{LLM orchestration.}
A large body of work studies how to trade off LLM performance and inference cost by adaptively routing each query to a pool of models with different sizes or prices. Cascade-style methods such as FrugalGPT~\citep{chen2023frugalgpt} begin with a cheap model and escalate to stronger ones only when necessary, typically using confidence or other heuristic signals. More recent approaches learn query-level routers more directly, including difficulty-aware routing in HybridLLM~\citep{ding2024hybridllm}, preference-trained routing in RouteLLM~\citep{ong2024routellm}, and meta-model-based formulations such as FORC~\citep{vsakota2024fly}. \citet{si2025ccpo} formulated cost-effective LLM routing as a constrained Markov decision process, where a conformal policy is trained to optimize efficiency while satisfying target coverage requirements. Despite their differences, these methods operate at the \emph{query} level: a single routing decision selects the model for the entire response, and the router typically relies on prompt-level or response-level signals rather than intermediate chain-of-thought states. As a result, they are not designed to exploit the evolving dependencies along a multi-step reasoning trajectory. 

\textbf{Stepwise routing and process-level collaboration.} Among the methods most relevant to our setting are stepwise routing approaches, which model multi-step reasoning as a sequential decision process over intermediate chain-of-thought states. RSD~\citep{liao2025reward} and SpecReason~\citep{pan2025specreason} use step-level reward assessments from either a trained process reward model or an LLM judge to decide whether a draft step generated by a smaller model should be accepted or re-generated by a larger model. STEER~\citep{lee2025steer} shows that logit-based confidence, when calibrated through a stepwise Gaussian mixture model, can act as a lightweight routing signal for per-step reasoning guidance. Post-hoc conformal prediction has also been used for LLM decision-making in stepwise planning tasks, for example in Robots that Ask for Help~\citep{ren2023robots}. All of these methods require post hoc threshold selection, which is typically carried out through extensive grid search.

\begin{figure}[t]
    \centering
    \includegraphics[width=0.98\linewidth]{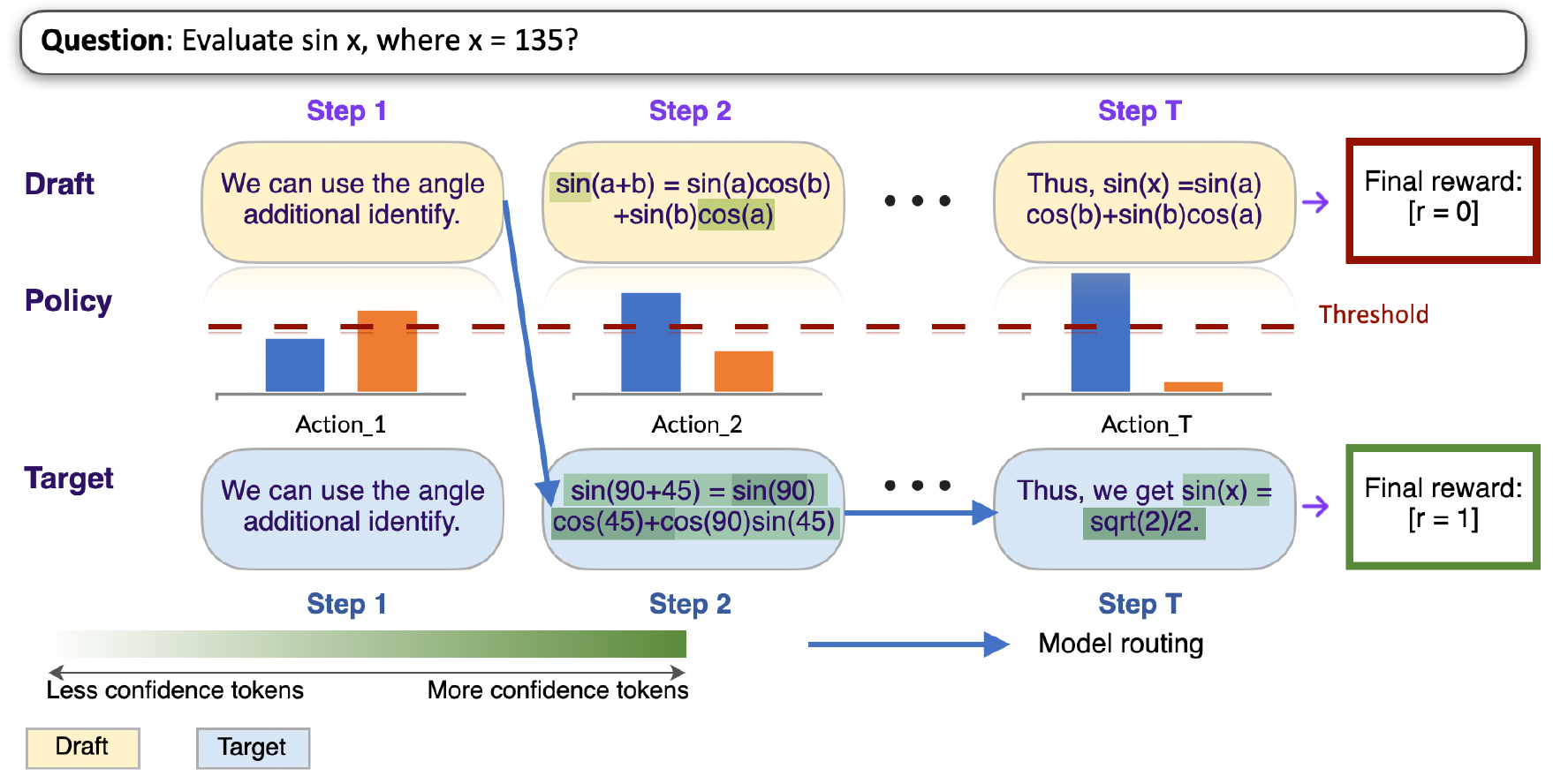}
    \caption{Illustrations of policy-guided stepwise model routing.}
    \label{fig:placeholder}
\end{figure}

\section{Problem Setting}

In this section, we formulate guided stepwise model routing as a constrained Markov decision process. Specifically, we consider a question-answering task in which question-answer pairs $(x, y^*)$ are drawn from an underlying distribution $\mathcal{D}$.

\subsection{Stepwise Model Routing}

We consider a pool of $K$ large language models $(M_1,\dots,M_K)$, ordered by increasing capability and cost; while our implementation and experiments focus on $K=2$, our algorithm is general. For a given input problem $x$, the system constructs a chain-of-thought reasoning trajectory over a finite horizon of $T$ steps. At step $t$, we denote the partial trajectory by
\[
    h_t = (x, y_{1:t}),
\]
where $y_{1:t}$ represents the first $t$ CoT steps generated so far.

We formalize the routing process as a finite-horizon Partially Observable Markov Decision Process (POMDP). At step $t=0$, we first generate the initial reasoning step $y_0$ using the base model. Then, for each step $t=1,\dots,T$, we collect the next reasoning step from the model selected by action $a_t$. The policy $\pi$ receives an observation $o_t \in \mathcal{O}$, where $o_t$ consists of black-box information derived from $h_t$. The policy also tracks the current model $M_k$. Based on the latest observation, the policy selects an action
\[
    a_t \in \mathcal{A} = \{\text{continue with } M_k\}\cup\{\text{escalate to } M_{k'} \mid k'>k\}.
\]
The selected model then generates the next reasoning steps, and the reasoning trajectory continues accordingly. Each reasoning step incurs an inference cost
\begin{equation}\label{eq:c}
    c(h_{t-1}, a_t, y_t),
\end{equation}
for example, approximate FLOPs or API cost. This cost may depend not only on the chosen model, but also on the generation history, since API pricing can depend on the input context as well. After termination, the system receives a task-specific reward
\begin{equation}\label{eq:r}
    r(x, \hat{y}, y^\star),
\end{equation}
where $y^\star$ is the ground-truth answer and $\hat{y}$ is the parsed answer extracted from the final reasoning step $y_T$. The total reward and total cost for problem $x$ are therefore given by
\[
    R(x) = r(x, \hat{y}, y^\star), \qquad
    C(x) = \sum_{t=1}^T c(h_{t-1}, a_t, y_t).
\]
Intuitively, routing to the larger model at any given step increases the expected reward. However, in the setting of cost-efficient reasoning, such routing decisions also incur additional computational cost and may therefore be unnecessary.

\subsection{Policy Learning Problem}

Our goal is to learn a stepwise routing policy that decides, at each intermediate reasoning state, whether to continue with the current smaller model or escalate to a stronger one. For clarity, we first consider the two-model setting consisting of a smaller draft model $m$ and a larger target model $M$. Let $\pi(a_t \mid h_t)$ denote a routing policy over actions $a_t \in \mathcal{A}$. Executing $\pi$ induces a stopping time $T_{\pi}$ and a final parsed answer $\hat{y}_{\pi}$. We define the total reward and total cost under $\pi$ by
\[
R_\pi(x) := r(x,\hat{y}_{\pi},y^\star),
\qquad
C_\pi(x) := \sum_{t=1}^{T_{\pi}} c(h^{\pi}_{t-1},a_t,y^{\pi}_t).
\]
In our exact setting, the terminal reward is binary and indicates answer correctness, so that $R_\pi(x)\in\{0,1\}$. Our objective is to reduce inference cost while preserving the correctness level attained by the stronger model. Accordingly, an effective routing policy should continue with the cheaper model whenever it is likely to succeed, and escalate only when the smaller model is unlikely to maintain sufficient answer quality. To formalize this requirement, we consider a threshold policy
\begin{align*}
\Pi_{\pi,\kappa}(h) = \{a \in \mathcal{A} : \pi(a \mid h) \ge \kappa\},
\end{align*}
where $\kappa$ is the threshold. Given a target error level $\alpha \in (0,1)$, we seek a policy $\Pi$ that minimizes the expected inference cost subject to a relative correctness constraint:
\begin{equation}\label{eq:problem}
\min_{\Pi} \mathbb{E}\big[C_\Pi(x)\big]
\qquad
\text{s.t.}
\qquad
\mathbb{P}\big(R_\Pi(x)=1 \lor R_M(x)=0\big) \ge 1-\alpha.
\end{equation}
Equivalently, with probability at least $1-\alpha$, the routed system must be correct whenever the stronger model $M$ is correct. Among all policies satisfying this constraint, we prefer the one with the smallest expected computation cost.

\section{Threshold Constrained Policy Optimization for Routing}

We parameterize the base routing policy $\pi$ by a simple neural network that maps $o_t$ to a distribution over actions. We jointly optimize $\pi$ and $\kappa$ to solve \eqref{eq:problem}. We describe how to optimize $\pi$ in Section~\ref{sec:policylearning} and how to optimize $\kappa$ in Section~\ref{sec:thresholdcalibration}.

\subsection{Threshold Policy Learning}
\label{sec:policylearning}

We first estimate the value function for the threshold-filtered actions on the stochastic policy $
S_{\pi,\kappa}(a \mid o)
=
\frac{\mathbf{1}\{\pi(a \mid o)\ge \kappa\}}
{\sum_{a' \in \mathcal{A}} \mathbf{1}\{\pi(a' \mid o)\ge \kappa\}}$. To bridge the gap, we employ V-trace off-policy correction~\citep{espeholt2018impala,si2025ccpo}. At each iteration, rollouts under $\pi$ produce tuples $(o_t,a_t,r_t,c_t)$, where $r_t$ and $c_t$ are the reward and constraint values on step $t$ (\eqref{eq:r} and \eqref{eq:c}), respectively. 
Then, the V‐trace target is optimized via gradient descent on both the reward and constraint function estimates.

Next, we apply the trust–region policy update. We iteratively solve the following approximation of~\eqref{eq:problem} (we describe how $\kappa$ is updated later):
\begin{align}
\label{eq:policyupdate}
\min_{\Pi} \mathds{E}\left[\sum_{t=1}^T \hat{A}^{\Pi_{\pi,\kappa}}_t\right]
\qquad\text{s.t.}\qquad
\bar{J}_{C}^{\Pi,\kappa}\ge 1 - \alpha,
\quad D_{\mathrm{KL}}(S_{\pi, \kappa}\|S_k)\le\delta.
\end{align}

Here, the KL constraint keeps the new policy close to the current one, and $\bar{J}_C^{\Pi,\kappa}$ is an approximation of the constraint value:
\begin{align*}
\bar{J}_C^{\Pi,\kappa}&\approx\mathbb{P}\big(R_\Pi(x)=1 \lor R_M(x)=0\big)
\le\Pr[R_\Pi(x)=1]+\Pr[R_M(x)=0].
\end{align*}
The second term is easy to estimate; for the first, we have 
\begin{align*}
\Pr[R_\Pi(x)=1] 
\approx\mathds{E}_\pi\left[\left(\prod_{t=1}^T\rho_t\right)\mathds{1}\{R_\pi(x)=1\}\right]
\quad\text{where}\quad
\rho_t= \min\left\{ \bar{\rho}, \frac{\Pi_{\pi, \kappa}(a_t \mid o_t)}{\pi(a_t \mid o_t)} \right\}.
\end{align*}
Here, the clipping value is $\bar{\rho} = 2$. The approximation arises from clipping the importance weights $\rho_t$. As in CPO~\citep{achiam2017constrained}, we then compute the optimal Lagrange multiplier for the constraint and use conjugate gradient descent to solve for the natural gradient step that satisfies the KL radius $\delta$; the resulting step gives the new policy $\pi_{k+1}$.

\subsection{Threshold Calibration}
\label{sec:thresholdcalibration}

We follow the threshold update rule for $\kappa$~\citep{angelopoulos2024online, si2025ccpo}. Concretely, when solving for $\pi_{k+1}$ in~\eqref{eq:policyupdate}, we fix $\kappa=\kappa_k$. Then, we update $\kappa_k$ according to the coverage of $\Pi_{{k+1},\kappa_k}$: 
\begin{equation}\label{eq:ccpo_decay}
 \kappa_{k+1}
 =
 \kappa_k + \eta_k \left(1 - \mathds{1}\{R_{\Pi_{{k+1},\kappa_k}}(x)=1 \lor R_M(x)=0\} - \alpha\right).
\end{equation}
Under the standard step-size conditions
$
\sum_t \eta_t = \infty
$
and
$
\sum_t \eta_t^2 < \infty,
$ this update maintains the coverage with statistical guarantee. 
We adapt this algorithm (which was originally proposed for conformal prediction) to stepwise routing. The key challenge is that the quality of each sampled intermediate behavior is context-dependent and generally cannot be measured exactly. A fully faithful solution would require exhaustive trajectory-level verification over the induced reasoning tree, which is computationally intractable in practice. We therefore propose an efficient approximation based on local verification.

Concretely, we introduce an auxiliary per-step verifier LLM $V$ that produces a binary judgment  $V(o_t,\tilde{y}_t)\in\{0,1\}$ for only the reasoning step selected at the current state. When the selected step is judged incorrect, we check whether the ``escalate" action is contained in the current thresholded action set. The trajectory is locally verified if $\forall t \in {1,\dots,T_\pi}; V(o_t,\tilde y_t)=1$. Thus, the update rule of the threshold in~\eqref{eq:ccpo_decay} becomes 
\begin{equation*}
 \kappa_k + \eta_k \left(1 - \mathds{1}\{R_\pi(x)=1 \lor R_M(x)=0 \lor \prod_{t=1}^{T_\pi} V(o_t,\tilde{y}_t)=1\} - \alpha\right).
\end{equation*}

This yields a practical surrogate for estimating both coverage and the action-set-size constraint during policy training. Since the verifier is used only during training, the approach does not incur any additional inference-time overhead.

Intuitively, the threshold provides a conservative regularizing effect for RL training under highly stochastic uncertainty signals. At inference time, we find that when the action set contains both options, executing only the larger model is sufficient to provide strong performance as it corresponds to the action with the highest value, making it equivalent to 
\begin{equation}\label{eq:thresh}
    \Gamma(o_t) =\begin{cases}
1, & \text{if } \pi( \text{escalate} \mid o_t) \ge \kappa,\\
0, & \text{otherwise.}
\end{cases}
\end{equation}

\subsection{Confidence Estimation}

In this section, we describe how $o_t$ is constructed from $h_t$, with particular emphasis on the uncertainty scores. For each reasoning step $i$, let $y_i = (y_{i,1}, \dots, y_{i,L_i})$ be the generated \textit{token} sequence and let $\mathcal{M}_i$ denote the set of math-only token positions. In the open-only setting, where full logits are accessible, we follow \citet{lee2025steer} and define
\[
S^{\mathrm{os}}_i
:=
\frac{1}{|\mathcal{M}_i|}
\sum_{j \in \mathcal{M}_i}
\max_{v \in \mathcal{V}} z_{i,j,v},
\]
where $z_{i,j,v}$ is the pre-softmax logit for vocabulary token $v$ at position $j$. 

In the API setting, only top-$5$ log-probabilities are available. Let $\tilde p_{i,j}$ be the top-1 probability of the generated token $y_{i,j}$ included in the returned top-$5$ set. We then define
\[
S^{\mathrm{API}}_i
:=
\min_{j \in \mathcal{M}_i^{(5)}} \tilde p_{i,j}.
\]
Thus, the open-only score measures average logit-based confidence over mathematically salient tokens, while the API score uses the weakest observed math-token probability as an alternative under limited access to model internals. Given the flexibility of the policy-based framework, we also augment the covariates with several simple forms of black-box information. In addition to uncertainty-related signals, we include the current step index, an optional problem-difficulty indicator, whether the current step corresponds to the final answer, and the length of the current reasoning step. These features are lightweight and broadly available, yet they provide useful contextual information for routing decisions. 

\section{Experiments}

\subsection{Experimental Setup}

\textbf{Benchmarks.}
We evaluate our method on three mathematical reasoning benchmarks:
\begin{itemize}
\item GSM8K~\citep{cobbe2021training}: a benchmark of 8.5K grade-school math word problems that require multi-step arithmetic reasoning in natural language.
\item MATH500~\citep{hendrycks2021measuring, lightman2023let}: a 500-problem subset of the MATH benchmark, commonly used to evaluate mathematical reasoning.
\item OmniMath~\citep{gao2024omni}: an Olympiad-level mathematics benchmark covering a diverse range of mathematical subdomains and difficulty levels.
\end{itemize}

For each dataset, we prompt the models to generate step-by-step chain-of-thought reasoning by specifying stop words and logging intermediate states and reasoning steps. We split $\mathcal{D}$ into a training set $\mathcal{D}_{\text{train}}$ and a calibration set $\mathcal{D}_{\text{cal}}$.  
We provide the prompt in Appendix~\ref{app:prompt}. 

\textbf{Answer evaluation.}
Accurate answer evaluation is critical for our policy training. Because the training objective is tied to a distribution-free coverage constraint, noisy correctness labels can directly distort the coverage signal; this effect is further amplified by the stochasticity of small-batch training, often leading to overly conservative policies. In mathematical reasoning, this challenge becomes pronounced due to heavy LaTeX formatting, gibberish generations, parsing failures, and equivalent answers expressed in different forms. To address this issue, we incorporate xVerify~\citep{xVerify} during training. xVerify is a verifier tailored to reasoning-model evaluation that can extract final answers from long outputs and identify equivalence across diverse mathematical and textual forms, thereby providing a substantially more reliable correctness signal for policy optimization.

\textbf{Models.} We conduct experiments in two settings. First, in the open-only setting, we evaluate the \textit{Qwen2.5-Math-Instruct} models~\citep{yang2024qwen2} optimized for mathematical reasoning, using the 1.5B model as the small draft model and the 7B model as the large target model. Second, in the open-to-closed setting, we use a Qwen model as the small draft model and GPT-4.1-mini as the large target model. GPT-4o-mini was used as the stepwise coverage checking model. For the RSD baseline~\citep{liao2025reward}, we use \textit{Skywork-o1-PRM-1.5B}~\citep{he_2024_16998085}. Note that the \textit{Skywork-o1-PRM} models are trained on the \textit{Qwen2.5-Math-Instruct} models. All experiments were conducted using vLLM~\citep{kwon2023efficient} with a temperature of 0.7. We define a reasoning step as a text segment separated by \texttt{\textbackslash n\textbackslash n}.

\textbf{Baselines.} We adopt two groups of stepwise routing baselines: \textit{External models} and \textit{No external models}. For the \textit{External models} baselines, we adopt RSD~\citep{liao2025reward}. For the \textit{No external models} baselines, we adopt (1) the \textit{SpecReason}~\citep{pan2025specreason} design and (2) the very recent stepwise reasoning framework STEER~\citep{lee2025steer}.

\begin{itemize}
\item SpecReason~\citep{pan2025specreason} uses a larger LLM as a judge to generate verbalized scores, and then thresholds them to produce step-level rewards.
\item RSD~\citep{liao2025reward}: In reward-guided speculative decoding (RSD), each reasoning step is initially produced by the draft model $m$. A process reward model (PRM) then assigns a scalar reward to the generated step, and the target model $M$ is used to regenerate the step only if the reward is below a predefined threshold.

\item STEER~\citep{lee2025steer} utilizes the LLM's logit-based uncertainty score for stepwise routing. As the scores are inherently highly stochastic, it uses stepwise GMMs to determine whether to switch to the large model for subsequent steps. Like other works, they threshold these decisions. 
\end{itemize}

For all baselines, we follow their original implementation that performs a grid search over threshold values with a step size of 0.1, selecting the best value for fair comparison. For our method, we set a small target error rate $\alpha = 0.02$. For threshold policy training, we use policy networks that are three-layer neural networks with 64 hidden units per layer. We train them from scratch for 1000 steps with a learning rate of $10^{-3}$ and batch size of 5. We set the threshold update step size to 0.1. For API-cost experiments with GPT guidance, we adapt the baselines to be compatible with OpenAI APIs and prompt formatting.

\textbf{Metrics.}
We report the following metrics:
\begin{itemize}
\item \textbf{Accuracy}: The fraction of problems solved correctly.
\item \textbf{Average cost}: The average cumulative inference cost per problem, measured either in normalized FLOPs or in API monetary cost. Following prior work~\citep{liao2025reward,sardana2023beyond,kaplan2020scaling}, we estimate open Transformer inference cost using the standard approximation of $2N$ FLOPs per generated token for a model with $N$ parameters. We report the average FLOPs consumed per query, with FLOPs presented in units of $10^{12}$ for readability. The API cost is the average monetary cost per problem, including both input and output tokens. Since these tokens are billed at different rates, we account for them separately, and the cost is presented in units of US cents. For GPT-4.1-mini, the price is \$0.40 per 1M input tokens, \$0.10 per 1M cached input tokens, and \$1.60 per 1M output tokens.
\item \textbf{Accuracy-per-Cost (A/C)}: Following \citet{ma2024kor}, we also report Accuracy-per-Cost to better capture the trade-off between reasoning performance and inference cost. Note that, in the API cost setting, this metric may only be meaningful when the compared methods achieve similar accuracy levels.
\end{itemize}

\subsection{Results}

Tables~\ref{tab:flop} and \ref{tab:api} summarize the main quantitative results. In the open-only setting, we achieve a strong accuracy-cost trade-off across all three mathematical reasoning benchmarks. On GSM8K, our method reaches $94.5$ accuracy with $2.03$ FLOPs, essentially matching the 7B-only model ($94.6$) while using less than half of the computation. On MATH500, we attain the highest overall accuracy among the compared methods while remaining substantially cheaper than the 7B-only baseline. On OmniMath, our method remains competitive with RSD, the strongest stepwise baselines with an external reward model. Overall, these results suggest that threshold policy-guided routing is able to recover the large model's reasoning ability while materially reducing inference cost.

\begin{table}[t]
\centering
\setlength{\tabcolsep}{2.1pt}
\begin{tabular}{l|c|ccc|ccc|ccc}
\hline
Method      & \multicolumn{1}{c|}{\multirow{2}{*}{\begin{tabular}[c]{@{}c@{}}Ext.\\ Models\end{tabular}}} & \multicolumn{3}{c|}{GSM8K}                        & \multicolumn{3}{c|}{MATH500}                      & \multicolumn{3}{c}{OmniMath}                      \\
&                   & Acc$\uparrow$ & FLOPs$\downarrow$ & A/C$\uparrow$ & Acc$\uparrow$ & FLOPs$\downarrow$ & A/C$\uparrow$ & Acc$\uparrow$ & FLOPs$\downarrow$ & A/C$\uparrow$ \\ \hline
1.5B Only   &    --       & 85.2          & 0.973             & 87.56         & 73.0          & 1.82              & 40.1          & 26.4          & 2.72              & 9.7           \\
7B Only     & --        & 94.6          & 4.42              & 21.40         & 79.6          & 9.10              & 8.75          & 28.2          & 13.56             & 2.08          \\ \hline
SpecReason  &   $\times$       & 85.2          & 1.03              & 82.72         & 77.4          & 7.74              & 10.0          & 28.0          & 12.8              & 2.18          \\
RSD         &     $\checkmark$  & 94.4          & 2.27              & 41.59         & 79.8          & 4.56              & 17.5          & 30.0          & 8.54              & 3.51          \\
STEER       &  $\times$   & 94.4          & 2.66              & 35.49         & 79.6          & 6.38              & 12.4          & 28.3          & 9.16              & 3.09          \\
Ours &   $\times$    & 94.5          & 2.03              & 46.55         & 82.8          & 5.34              & 15.5          & 29.1          & 8.24              & 3.53          \\ \hline
\end{tabular}
\caption{Representative performance on GSM8K, MATH500, and OmniMath measured by FLOPs. Both the source and target models are open Qwen models.}
\label{tab:flop}
\end{table}

The open-to-closed setting presents a somewhat more mixed picture, due to limitations of the OpenAI API log-probability scores and formatting differences between the Qwen and GPT models, particularly on the math-symbol-heavy OmniMath dataset. First, relative to directly calling GPT-4.1-mini, our method reduces mean API cost on GSM8K and MATH500 while maintaining comparable accuracy, and it also improves markedly over SpecReason in both cost and overall balance. Compared with STEER, we are more favorable on GSM8K and MATH500, where it attains similar or higher accuracy at lower API cost; on OmniMath, our method improves accuracy, while requiring lower cost than the. Relative to the external method RSD, the comparison is mixed: we have slightly higher accuracy on GSM8K, but RSD remains stronger on MATH500 and slightly stronger on OmniMath. While the results verify our advantage in effectiveness over all methods without external guidance, the larger gap compared to the PRM-based method RSD is possibly caused by the lower quality of the uncertainty scores we can achieve from the API, which only provides top-5 log-probabilities.

\subsection{Discussion}

\textbf{Our method captures the stepwise dependency.} An important advantage of our method is its ability to better capture stepwise dependency. Stepwise reasoning is naturally a sequential decision problem: whether a given step should remain with the small model or be escalated to the large model depends not only on the uncertainty at the current step, but also on the preceding trajectory and the future value of the routing decision. As a result, capturing inter-step dependency is important for achieving a strong accuracy-cost trade-off. 

Existing baselines address this aspect only partially. RSD measures the semantic quality of intermediate reasoning steps in a general sense, but does not explicitly model the comparative advantage between the small and large models at each step. SpecReason captures dependency more directly by conditioning the scoring model on the full prefix history, but this design also incurs substantial input-side inference overhead, particularly in API settings. STEER handles noisy confidence scores through GMM-based calibration, but its modeling is applied independently at each step, which limits its ability to capture dependency within a reasoning trajectory. In addition, these methods require post-hoc threshold tuning through grid search, which is inefficient and may restrict the granularity of the final chosen threshold. By contrast, our method integrates trajectory-level dependency into a sequential policy optimization framework and adaptively updates its routing boundary during training, leading to a more accurate and efficient stepwise routing strategy.

\textbf{Routing overhead and practical efficiency.} From a systems perspective, we also have an appealing operational advantage. SpecReason and RSD obtain routing signals through an additional LLM judge or a separately trained PRM, which introduces extra model evaluations and latency. By contrast, confidence-based routing methods like STEER avoid this dependency on external scoring at inference time, providing a latency advantage because of this design choice. Our method inherits much of this practical benefit while additionally removing the need for post-hoc threshold grid search. Thus, lightweight guidance methods like ours should be particularly attractive in deployments where the overhead of external reward evaluation is a primary bottleneck.

\begin{table}[t]
\centering
\setlength{\tabcolsep}{2.1pt}
\begin{tabular}{l|c|ccc|ccc|ccc}
\hline
Method       & \multicolumn{1}{c|}{\multirow{2}{*}{\begin{tabular}[c]{@{}c@{}}Ext.\\ Models\end{tabular}}} & \multicolumn{3}{c|}{GSM8K}                       & \multicolumn{3}{c|}{MATH500}                     & \multicolumn{3}{c}{OmniMath}                             \\
             & \multicolumn{1}{c|}{}                                                                       & Acc$\uparrow$ & Cost$\downarrow$ & A/C$\uparrow$ & Acc$\uparrow$ & Cost$\downarrow$ & A/C$\uparrow$ & Acc$\uparrow$ & Cost$\downarrow$ & $A$/C$\uparrow$ \\ \hline
7B Only      &   --   & $85.2$\tablefootnote{Since Qwen2.5-Math-Instruct-7B and GPT-4.1-mini exhibit the same accuracy on GSM8K, we restrict the open draft model on this dataset to Qwen2.5-Math-Instruct-1.5B.}      & 0.0              & --            & 79.6          & 0.0              & --            & 28.2          & 0.0              & --                    \\ 
GPT-4.1-mini &   --  & 95.2          & 0.0293           & 32.4          & 84.8          & 0.1313           & 6.46          & 43.0          & 0.3356           & 1.28                  \\ \hline
SpecReason   &  $\times$  & 94.9          & 0.0994           & 9.55          & 80.8          & 0.3653           & 2.21          & 26.0          & 0.5691           & 0.46                  \\ 
RSD          &   $\checkmark$  & 93.5          & 0.0093           & 100.5         & 87.1          & 0.0331           & 26.31         & 40.2          & 0.2932           & 1.52                  \\ 
STEER        &   $\times$ & 94.5          & 0.0608           & 15.54         & 84.0          & 0.0827           & 10.16         & 30.0          & 0.0823           & 3.64                  \\ 
Ours        &  $\times$  & 94.5          & 0.0205           & 46.09         & 85.0          & 0.0493           & 17.24         & 38.6          & 0.285            & 1.35                  \\ \hline
\end{tabular}
\caption{Performance on GSM8K, MATH500, and OmniMath in terms of API cost. The source model is an open Qwen model, and the target models are GPT models.}
    \label{tab:api}
\end{table}

\textbf{Discussion on the choice of uncertainty score.} The effectiveness of our method also supports the broader view that model-derived confidence contains useful information for stepwise orchestration. This view is well recognized and has been supported in adjacent settings such as speculative decoding~\citep{leviathan2023fast, chen2023accelerating} and MCMC-based chain-of-thought sampling~\citep{karan2025reasoning}. In our setting, this perspective helps explain why relatively lightweight uncertainty information, combined with simple contextual features, can still support a high-quality routing policy. The mixed API results, however, also suggest that raw log-probability scores may be less informative than logit-based confidence scores for stepwise mathematical reasoning problems. This is consistent with \citet{lee2025steer}, which shows that logit-based confidence is particularly effective in ambiguous or borderline cases, where simpler uncertainty surrogates become less discriminative.

\section{Conclusion}

We have presented a threshold policy training framework for guided stepwise model routing in cost-efficient reasoning. By formulating routing as a constrained sequential decision problem and combining policy optimization
, our method offers a principled alternative to methods based on verbalized rewards or stepwise GMMs with manual threshold tuning, while also serving as a lightweight and competitive alternative to methods that rely on external models. Across GSM8K, MATH500, and OmniMath, our method delivers strong accuracy-cost trade-offs, with especially favorable results in the open-only setting and competitive performance in the open-to-closed setting. More broadly, the results suggest that our method is promising for cost-aware reasoning systems that must balance performance, computation, and statistical reliability within a unified framework. Important directions for future work include direct latency evaluation, broader non-mathematical benchmarks, and richer uncertainty representations for multi-model routing.



\bibliography{colm2026_conference}
\bibliographystyle{colm2026_conference}

\appendix
\section{Prompt Templates}\label{app:prompt}

\begin{itemize}
    \item Prompt template for reasoning generation

    \begin{lstlisting}[
  backgroundcolor=\color{gray!10},
  breaklines=true,
  breakatwhitespace=false,
  columns=fullflexible
]
<|im_start|>system
Please reason step by step, and put your final answer within \boxed{}.
<|im_end|>

<|im_start|>user
{input}
<|im_end|>

<|im_start|>assistant
{output}
\end{lstlisting}

\item Prompt template for GPT stepwise verification

\begin{lstlisting}[
  backgroundcolor=\color{gray!10},
  breaklines=true,
  breakatwhitespace=false,
  columns=fullflexible,
  % basicstyle=\ttfamily\small
]
Task: Determine whether the final step is correct for the given problem.

Instructions:
- Reply with exactly one verdict: correct or incorrect.
- Do not explain, justify, or provide any reasoning.

Question:
{question}

Previous steps:
{prefix}

Last step:
{answer}

Verdict:
correct | incorrect
\end{lstlisting}
\end{itemize}

\end{document}